\documentclass[letterpaper]{article} % DO NOT CHANGE THIS
\usepackage{aaai25}  % DO NOT CHANGE THIS
\usepackage{times}  % DO NOT CHANGE THIS
\usepackage{helvet}  % DO NOT CHANGE THIS
\usepackage{courier}  % DO NOT CHANGE THIS
\usepackage[hyphens]{url}  % DO NOT CHANGE THIS
\usepackage{graphicx} % DO NOT CHANGE THIS
\usepackage{natbib}  % DO NOT CHANGE THIS AND DO NOT ADD ANY OPTIONS TO IT
\usepackage{caption} % DO NOT CHANGE THIS AND DO NOT ADD ANY OPTIONS TO IT
\usepackage{algorithm}
\usepackage{algorithmic}
\usepackage{subfigure}
\usepackage{newfloat}
\usepackage{booktabs}
\usepackage{listings}
\usepackage{makecell}
\usepackage{amsmath,amssymb}

\DeclareCaptionStyle{ruled}{labelfont=normalfont,labelsep=colon,strut=off} % DO NOT CHANGE THIS
\floatstyle{ruled}
\newfloat{listing}{tb}{lst}{}
\floatname{listing}{Listing}

\usepackage{xcolor}
\usepackage{xspace}

\newcommand{\ie}{{\emph{i.e.}}\xspace}

\newcommand{\etal}{{\emph{et al.}}}
\newcommand{\name}{FlexiTex}
\title{FlexiTex: Enhancing Texture Generation via Visual Guidance}
\author{
    Dadong Jiang\textsuperscript{\rm 1,2},
    Xianghui Yang\textsuperscript{\rm 2},
    Zibo Zhao\textsuperscript{\rm 2}, 
    Sheng Zhang\textsuperscript{\rm 2},
    Jiaao Yu\textsuperscript{\rm 2},
    Zeqiang Lai\textsuperscript{\rm 2}, \\
    Shaoxiong Yang\textsuperscript{\rm 2}, 
    Chunchao Guo\textsuperscript{\rm 2}\thanks{Corresponding author.}, 
    Xiaobo Zhou\textsuperscript{\rm 1}\thanks{Corresponding author.}, 
    Zhihui Ke\textsuperscript{\rm 1}
}

\affiliations{
    \textsuperscript{\rm 1}Tianjin University \\ 
    \textsuperscript{\rm 2}Tencent Hunyuan \\
    \{patrickdd, xiaobo.zhou\}@tju.edu.cn, \{seanxhyang, zibozhao, chunchaoguo\}@tencent.com
}

\usepackage{bibentry}
\begin{document}

\makeatletter
\let\@oldmaketitle\@maketitle%
\renewcommand{\@maketitle}{\@oldmaketitle%
 \centering
 \vspace{-8mm}
    \includegraphics[width=0.9\textwidth]{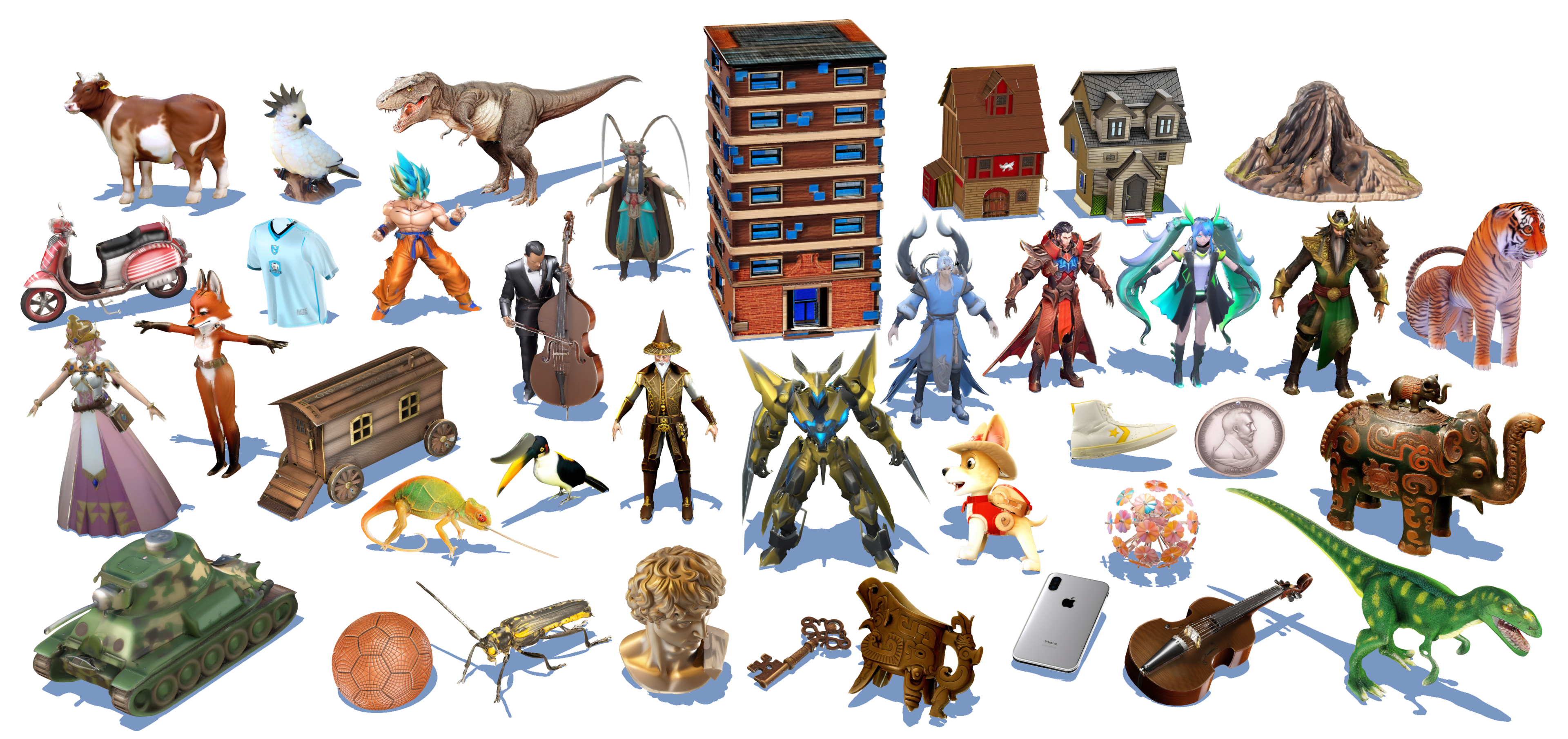}
     \captionof{figure}{A collection of textured results generated by \name. We can produce high-fidelity textures across various objects. }
    \label{fig: teaser}
    \bigskip}
\makeatother

\maketitle

\begin{abstract}
% 1. a big progress
% 2. drawbacks remain
% 3. observation / motivation
% 4. how to overcome
% 5. framework
% 6. module 1 & module 2
% 7. results
% 8. project page
% \zb{
Recent texture generation methods achieve impressive results due to the powerful generative prior they leverage from large-scale text-to-image diffusion models.
However, abstract textual prompts are limited in providing global textural or shape information, which results in the texture generation methods producing blurry or inconsistent patterns.
To tackle this, we present FlexiTex, embedding rich information via visual guidance to generate a high-quality texture.
The core of FlexiTex is the Visual Guidance Enhancement module, which incorporates more specific information from visual guidance to reduce ambiguity in the text prompt and preserve high-frequency details.
To further enhance the visual guidance, we introduce a Direction-Aware Adaptation module that automatically designs direction prompts based on different camera poses, avoiding the Janus problem and maintaining semantically global consistency.
Benefiting from the visual guidance, FlexiTex produces quantitatively and qualitatively sound results, demonstrating its potential to advance texture generation for real-world applications. 
% }
% Although recent texture generation methods have achieved impressive results by using text-to-image models, the issues of over-smoothness and inconsistency continue to pose challenges.
% To address these issues, we propose \name{}, a flexible framework for high-quality texture generation in this paper, supporting both text and image conditioning. The core of our work is the Visual Guidance Enhancement module, which reduces the ambiguities in text descriptions by incorporating more specific information from visual guidance. This approach prevents variance degradation and preserves high-frequency details when synchronizing the shared texture.
% Furthermore, we introduce a Direction-Aware Adaptation module that automatically designs direction prompts based on different camera poses, alleviating the Janus problem and maintaining semantically global consistency. \name{} outperforms existing methods quantitatively and qualitatively, showing its potential for real-world applications. 
% Project Page: \textcolor{blue}{\url{https://patrickddj.github.io/FlexiTex/}}
\end{abstract}

\begin{links}
    \link{Project}{https://patrickddj.github.io/FlexiTex/}
\end{links}

% \begin{figure*}
%     \centering
%     \includegraphics[width=0.9\textwidth]{images/banner.png}
%      \captionof{figure}{A collection of textured results generated by \name. We can produce high-fidelity textures across various objects. }
%     \label{fig: teaser}
%     \vspace{-5mm}
% \end{figure*}

\section{Introduction}

The field of computer graphics aims to enhance the interaction with 3D assets, but the scarcity of high-quality 3D assets is a significant challenge. The creation of such assets requires extensive artistic skills and is a labor-intensive process. However, recent advancements in deep generative models introduce a new paradigm known as Artificial Intelligence Generated Content (AIGC), revolutionizing the creation and manipulation of 3D models~\cite{hong2023lrm, mvedit2024}. Among the various aspects of asset generation, texture generation~\cite{chen2023text2tex, metzer2023latent} plays a crucial role in adding expressiveness to shapes, and finding applications in industries such as AR/VR, film, and gaming. Furthermore, texture generation holds the potential to overcome the limitations associated with collecting high-quality textured 3D assets.

Recent diffusion models show remarkable advancements in image generation. Noticing their potential on 3D tasks, researchers utilize them to generate textured 3D assets by optimization~\cite{metzer2023latent, poole2022dreamfusion} and inpainting~\cite{chen2023text2tex, zeng2024paint3d}, where they render 3D geometry into multiple views and update each view using rich 2D priors, However, these methods neglect cross-view correspondence in 3D space, leaving style inconsistency and seams. Recent works~\cite{zhang2024texpainter, liu2023text, liu2024vcd, gao2024genesistex} keep the style consistent and reduce the seams by utilizing batch inference and synchronize multi-view denoising on a shared texture, while still leaving heavy cross-view inconsistency. We argue that it is hard to keep cross-view alignment relying on ambiguous text prompts solely, and frequent feature aggregation introduces a serious variance bias~\cite{liu2024vcd}, resulting in over-smoothness. 
Moreover, they suffer from the \textbf{Janus problem}, due to lack of geometry-aware ability and the bias introduced by ambiguous text descriptions.

Noticing these, we propose a simple yet effective framework to achieve high-quality, precise texture generation, named ~\name, which supports text and image conditions both. \textbf{First}, we mitigate the ambiguity of text descriptions by converting them to more explicit modalities, \ie, image, which serves as a guide during batch inference and specifies the target object more accurately. Specifically, we introduce a Visual Guidance Enhancement module upon diffusion models to convert the text into an image and inject such a more specific informative guide into the batch inference process through cross-attention. By this, we ensure a more consistent direction during denoising and prevent variance degradation in joint sampling on the shared latent texture.
\textbf{Second}, we apply a Direction-Aware Adaptation module to alleviate the Janus problem, where we inject direction prompts under different views into the model. This module makes the model more sensitive to direction information and encourages semantic alignment between views.
\textbf{Third}, the proposed method is not only training-free but also more flexible than previous methods, as it accepts image prompts straightforwardly, by which it can finish texture transfer. This adaptability makes our framework suitable for various geometries and provides more diversity in the generated textures.
We conduct comprehensive studies and analyses involving numerous 3D objects from various sources to demonstrate the effectiveness of ~\name in texture generation, as in Fig.~\ref{fig: teaser}.

We summarize our contributions as follows: 
\begin{itemize}
    \item We present a novel and flexible framework for high-quality and accurate texture generation,~\name, which supports text-conditioned generation, image-conditioned generation, and texture transfer tasks. 
    \item We introduce the Visual Guidance Enhancement module to convert the ambiguous text descriptions into more specific images to specify the target between the plausible results.
    \item We design the Direction-Aware Adaptation module by providing explicit directional information to alleviate Janus problems, which encourages semantic alignment with contents under different views.
    % \item We conduct a comprehensive study involving numerous 3D objects from various sources. The experimental results demonstrate the superiority of our method over baseline methods.
\end{itemize}

\section{Related Work}
\noindent\textbf{Text-to-Image Diffusion Models. } 
Recent years have witnessed the advancements of text-to-image diffusion models~\cite{ramesh2022hierarchical, rombach2022high, saharia2022photorealistic, zhang2023adding, luo2023latent, casas2023smplitex} for the impressive generative capability to create high-fidelity images. 
Specifically, Stable Diffusion~\cite{rombach2022high} incorporates a text encoder from CLIP~\cite{radford2021learning} and generates realistic images based on input text prompts.
Beyond the text conditioning, Zhang~\etal~\cite{zhang2023adding} further enhances the model's capabilities. It allows the denoising network to be conditioned on additional input modalities, such as depth or normal maps. Furthermore, Ye~\etal~\cite{ye2023ip} design the effective and lightweight IP-Adaptor to achieve image prompt capability for the pre-trained text-to-image diffusion models. It can be inserted into the current framework and achieve multi-modal image generation. 
In our work, we leverage ControlNet and Stable Diffusion to offer geometrically conditioned image priors, and IP-Adaptor to support image guidance.

\noindent\textbf{Text-to-texture Synthesis.}
Generating textures on empty models is a challenging task, which requires precise alignment with geometry and semantic coherence. 
Early works~\cite{chen2022AUVNET, siddiqui2022texturify, yu2023texture, gao2022get3d, Chan2021, mitchel2024single} aim to utilize geometric prior for texture generation. 
They inject positional information and train geometry-aware generative models from scratch. However, these methods cannot generalize well on various categories and show blurry textures, due to the limited 3D data with high-quality textures for training.

Optimization-based methods~\cite{poole2022dreamfusion, metzer2023latent, pan2024enhancing, youwang2024paintit, chen2023fantasia3d, metzer2023latent, pan2024enhancing, zhang2024dreammat} utilize Score Distillation Sampling (SDS) on pre-trained diffusion models to update textures iteratively. 
% However, such methods are time-consuming and suffer from over-saturation.
Inpainting-based methods~\cite{chen2023text2tex, richardson2023texture, cao2023texfusion, zeng2024paint3d, zhang2023repaint123, tang2024intex, chen2023text2tex, cao2023texfusion, ahn2024contexture} design a sequential inpainting strategy, which fills blank areas of current views based on existing contents from neighbor views. 
Other methods~\cite{zhang2024clay, bensadoun2024meta, deng2024flashtex} finetune a multi-view diffusion model to generate a $2\times2$ gird for multi-view consistency, while they are trained on synthetic datasets and waste rich prior gained from large-scale real image datasets, showing poor diversity.

Recent studies focus on synchronization-based methods~\cite{liu2023text, liu2024vcd, gao2024genesistex, zhang2024texpainter, huo2024texgen}, as they claim that all views contribute equally to texture generation. SyncMVD~\cite{liu2023text} firstly uses a shared latent texture to force consistent latent features from multiple views during denoising.  
TexPainter~\cite{zhang2024texpainter} decodes all latent views and performs differentiable inverse rendering at each denoising step, aiming for a texture of higher resolution. 
GenesisTex~\cite{gao2024genesistex} maintains N unique textures with 1 shared texture to dynamically align the blending weight, and VCD-Texture~\cite{liu2024vcd} designs 3D-2D co-denoising to strengthen variance. 

However, optimization-based methods are time-consuming and suffer from over-saturation, and inpainting-based methods cannot maintain long-range consistency on the whole surface, due to the ambiguity of the text. Among synchronization-based methods, SyncMVD leads to disaturated or over-smoothed results, TexPainter introduces too much noise, and GenesisTex and VCD-Texture require significant memory and computational cost for cross-view attention. 
In contrast, our \name leverages visual guidance to generate textures, which not only excels in producing high-quality textures with rich details but also achieves fast and efficient generation.

\noindent\textbf{Image-to-texture Synthesis.}
Image-to-texture methods~\cite{richardson2023texture, zeng2024paint3d, pan2024enhancing, yeh2024texturedreamer, chen2023fantasia3d} take the image as the user input to generate textures. TEXTure~\cite{richardson2023texture} finetune DreamBooth LoRA~\cite{ruiz2022dreambooth} and uses the finetuned model as the updated base model for texture inpainting. 
Optimization-based methods PGC-3D~\cite{pan2024enhancing}, Fantastic3D~\cite{chen2023fantasia3d} and TextureDreamer~\cite{yeh2024texturedreamer} apply SDS conditioned on input image prompts for appearance modeling.
These methods require a long time for finetuning or repainting, and cannot transfer the semantic identity well to target meshes. 
Paint3D~\cite{zeng2024paint3d} applies image prompts directly on UV generation in the refinement stage, but cannot generalize well on complex UV maps, usually leaving certain artifacts on final textures.
Unlike these previous methods, our method can support high-quality texture generation in a training-free manner, which is flexible because it requires no additional adjustment for image prompts.

\begin{figure*}[tbp]
\centering
\includegraphics[width=\textwidth, trim=0 0 0 0, clip]{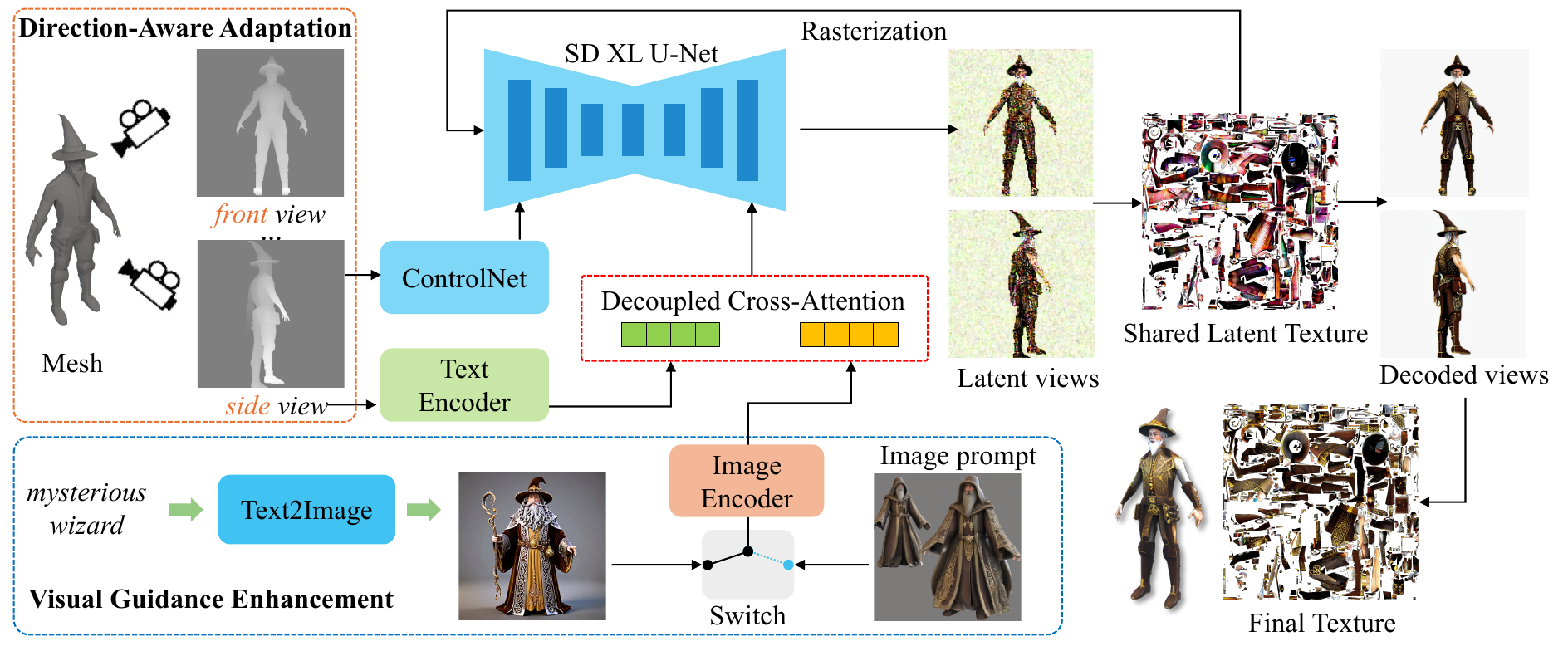} 
\caption{The framework of ~\name. Given a mesh with corrected orientation and text/image prompts, the Visual Guidance Enhancement module extracts \textbf{visual features} $\textcolor{green}{\blacksquare}$ to provide informative guidance during the denoising steps. The Direction-Aware Adaptation module then extracts \textbf{direction features} $\textcolor{orange}{\blacksquare}$ according to camera poses. Visual and direction features are integrated into the pre-trained UNet model with decoupled cross-attention, along with depth maps. Through iterative texture warping and rasterization, our method generates high-fidelity textures of both rich details and multi-view consistency. }
\label{fig: framework}
\end{figure*}

\section{Method}
\label{sec: method}

\name is designed to generate high-quality texture maps given an untextured mesh conditioned on either text or image. The overview is shown in Fig.~\ref{fig: framework}. In this section, we first provide the backgrounds of diffusion models, mesh rendering, and texture warping. Following this, we introduce the Visual Guidance Enhancement module, which is designed to overcome issues related to over-smoothing. Then, we present the Direction-Aware Adaptation module, which is specifically designed to incorporate direction information during texture generation, thereby addressing the Janus problem. 

\subsection{Preliminary}
\label{sec: preliminary}
% We greatly harness the generative prior, and introduce some preliminaries here.
We introduce the preliminaries here, including diffusion models, mesh rendering, and texture warping.

\noindent \textbf{Stable Diffusion \& Controlnet.} The diffusion model~\cite{ho2020denoising} consists of a forward process $q(\cdot)$ and a reverse denoising process $p_\theta(\cdot)$. The forward process progressively corrupts the original data, denoted as \( x_0 \),  with noise \( \epsilon \sim \mathcal{N}(0, 1) \) to a noisy sequence \( x_1, \ldots, x_T \), following a Markov chain as:
\begin{align}
\begin{split}
    q(x_t | x_{t-1}, y) &= \mathcal{N}(x_t; \sqrt{1 - \beta_t}x_{t-1}, \beta_t I),\\
    x_t &= \sqrt{\beta_t} x_0 + \sqrt{1 - \beta_t} \cdot \epsilon,
\end{split}
\end{align}
where \( \beta_t \) represents the variance schedule for \( t = 1, \ldots, T \); \( y \) is the condition.
The reverse process denoises the pure data from $x_T$ as:
\begin{equation}
    p_\theta(x_{t-1} | x_t, y) = \mathcal{N}(x_{t-1}; \mu_\theta(x_t, t, y), \Sigma_\theta(x_t, t, y)),
\end{equation}
where \( \mu_\theta \) and \( \Sigma_\theta \) denote the mean and variance predictions, respectively, which are obtained from a trainable network parameterized by $\theta$.

Besides, we introduce ControlNet~\cite{zhang2023adding} which injects low-level control, such as depth map $d$, during the denoising process of Stable Diffusion. The predicted noise of the U-Net with ControlNet is represented as $\boldsymbol{\epsilon}_{\theta}(x_t, y, d, t)$

\noindent \textbf{Texture warping}. 
Given a mesh $\mathcal{M}$, a texture map $\mathcal{T}$ and a viewpoint $C$, we can use the rasterization function $\mathcal{R}$ to render an image. 
% We do not consider anti-aliasing methods such as TriMap, to simplify the later process of back projection.
After rasterization, each valid pixel on rendered images corresponds to one on texture. However, these pixels are scattered in UV space after back projecting, necessitating Voronoi filling~\cite{voronoi} to fill up all blank regions in texture warping ($\mathcal{W}^{-1}$).
Given $\{\mathbf{s}_i | \mathbf{s}_i = (u_i, v_i)\}$ represents the UV coordinates of the $i$-th latent texel, we first generate the Voronoi diagram partitions the domain $\mathcal{T}$ into a set of regions $\{\mathcal{V}_i\}$, where each region $\mathcal{V}_i$ is defined as:
\begin{equation}
\mathcal{V}_i = \{\mathbf{p} \in \mathcal{T} \, | \, \|\mathbf{p} - \mathbf{s}_i\| \leq \|\mathbf{p} - \mathbf{s}_j\|, \forall j \neq i\}.
\end{equation}
Here, $\mathbf{p}$ is the 2D point position and $\|\cdot\|$ denotes the Euclidean distance.
Then we define a procedural function $\mathcal{P}_i: \mathcal{V}_i \to \mathbb{R}^3$ that generates the filling for the region $\mathcal{V}_i$ based on the texels $\mathbf{s}_i$ and the position $\mathbf{p}$ within the region.
Finally, we record visible masks on the texture map and apply specific boundary conditions near the edges to ensure a smooth and consistent filling.
The final texture is represented as the union of the filled regions: $\mathcal{T} = \bigcup\limits_{i=1}\limits^N V_i$, where $N$ is the number of regions.

\subsection{Visual Guidance Enhancement}
\label{sec: image_enhancement}

% Although previous methods force sampling on latent texture and synchronizing the batch inference, we find the generated results suffer from a great decline in texture diversity and quality (~\eg, over-smoothed details, low saturation). The ambiguity of the text modality, causes the batch diffusion process on multiple views to predict noise with large differences in directions. After being averaged on the latent texture, latent features become gradually over-smoothed in the intermediate process. Therefore, the final decoded image also has blurry details.

In~\name, we design a Visual Guidance Enhancement module to align the denoising inference on multiple views. Explicitly, for text input $\boldsymbol{x}^{text}$, we first utilize a text-to-image model to generate image prompt $\boldsymbol{x}_{img}$. Its corresponding semantic information $\boldsymbol{c}_{img}$ is injected into the denoising process in a cross-attention manner through IP-Adaptor~\cite{ye2023ip}. 
% IP-Adaptor designs a decoupled cross-attention mechanism where the cross-attention layers for text features and image features are separate. 
To be specific, given the image features $\boldsymbol{c}_{img}$, the output of new cross-attention $\mathbf{Z}'$ is computed as follows:
\begin{align}
    \begin{split}
    \mathbf{Z}' &=\text{Attention}(\mathbf{Q},\mathbf{K}',\mathbf{V}') \\
                &= \text{Softmax}(\frac{\mathbf{Q}(\mathbf{K}')^{\top}}{\sqrt{d}})\mathbf{V}',\\
    \end{split}
\end{align}
where, $\mathbf{Q}=\mathbf{Z}\mathbf{W}_q$, $\mathbf{K}'=\boldsymbol{c}_{img}\mathbf{W}'_k$ and $\mathbf{V}'=\boldsymbol{c}_{img}\mathbf{W}'_v$ are the query, key, and values matrices from the image features.
Since images provide more specific information, the image-guided batch diffusion process also benefits from this, in which case the denoising direction is more consistent. With intermediate latent features which are normalized by explicit visual features, there is no longer a huge variance degradation in joint sampling on shared latent texture, compared with the text-guided way. Images also provide clear guidance on appearance modeling, prompting a similar style on multiple views and maintaining global consistency.

In addition, our framework supports image-to-texture tasks. We do not require the image prompt to be totally aligned with mesh geometry but consider it a style-transferring task to maintain the most significant input semantic information on the target mesh.

\subsection{Direction-Aware Adaptation}
\label{sec: vaa}
% Despite improved details from visual guidance, we find the generated results have serious Janus problems. Actually, using text-to-image models in 3D generation continues to grapple with this significant issue, which arises from their bias towards front views and inadequate comprehension of 3D structures. This problem is characterized by the multiple appearances of a single anatomical feature (\eg, a face or an eye) at different positions on the object, leading to a reduction in the quality of the texture.
%  On one hand, we aim to solve Janus problems, keeping texture faithfulness, and on the other, image diversity from strong 2D prior of pre-trained text-to-image models should also be preserved.
%  Therefore, our question is: given an explicit mesh, can we directly turn a single-view generative model, which is more sensitive to geometry structures, into direction-aware generative models in a batch-inference manner, without additional training?

 We introduce Direction-Aware Adaptation which utilizes a Text Encoder with U-Net to provide spatial information and solve the Janus problem.
 We find that SD-XL is more sensitive to direction prompts. When we add prompts~\ie, \textit{front/side/back view}, it can understand these well and generate more semantically aligned results. Noticing that the orientation of the original shapes is not guaranteed, we first correct their orientation to ensure they are front-facing. This pre-processing step defines the standard orientation and determines the writing of subsequent direction prompts. Then we generate prompt $\{\boldsymbol{c}_{view, i}\}_{i=1}^{N}$ according to the elevation and azimuth for each view, organized as "from $<?>$ view".
Given such direction prompts, the denoising process for $view_{i}$ is conditioned on visual features $\boldsymbol{c}_{img}$ and direction features $\boldsymbol{c}_{view}$ via cross-attention, written as:

\begin{equation}
        \mathbf{Z}^{''}=\text{Softmax}(\frac{\mathbf{Q}\mathbf{K}^{\top}}{\sqrt{d}})\mathbf{V}+\text{Softmax}(\frac{\mathbf{Q}(\mathbf{K}')^{\top}}{\sqrt{d}})\mathbf{V}',\\
\end{equation}
where $\mathbf{Q}=\mathbf{Z}\mathbf{W}_q, \mathbf{K}=\boldsymbol{c}_{view}\mathbf{W}_k, \mathbf{V}=\boldsymbol{c}_{view}\mathbf{W}_v, \mathbf{K}'=\boldsymbol{c}_{img}\mathbf{W}'_k, \mathbf{V}'=\boldsymbol{c}_{img}\mathbf{W}'_v$. In our framework, the image prompt provides visual guidance on appearance and text description provides direction information, by which the direction-aware single-view generative model can naturally become a multi-view generator during batch inference and synchronized sampling, as shown in Fig.~\ref{fig: framework}.

\begin{figure*}[t!]
\centering
\includegraphics[width=0.9\textwidth, trim=0 20 0 25, clip]{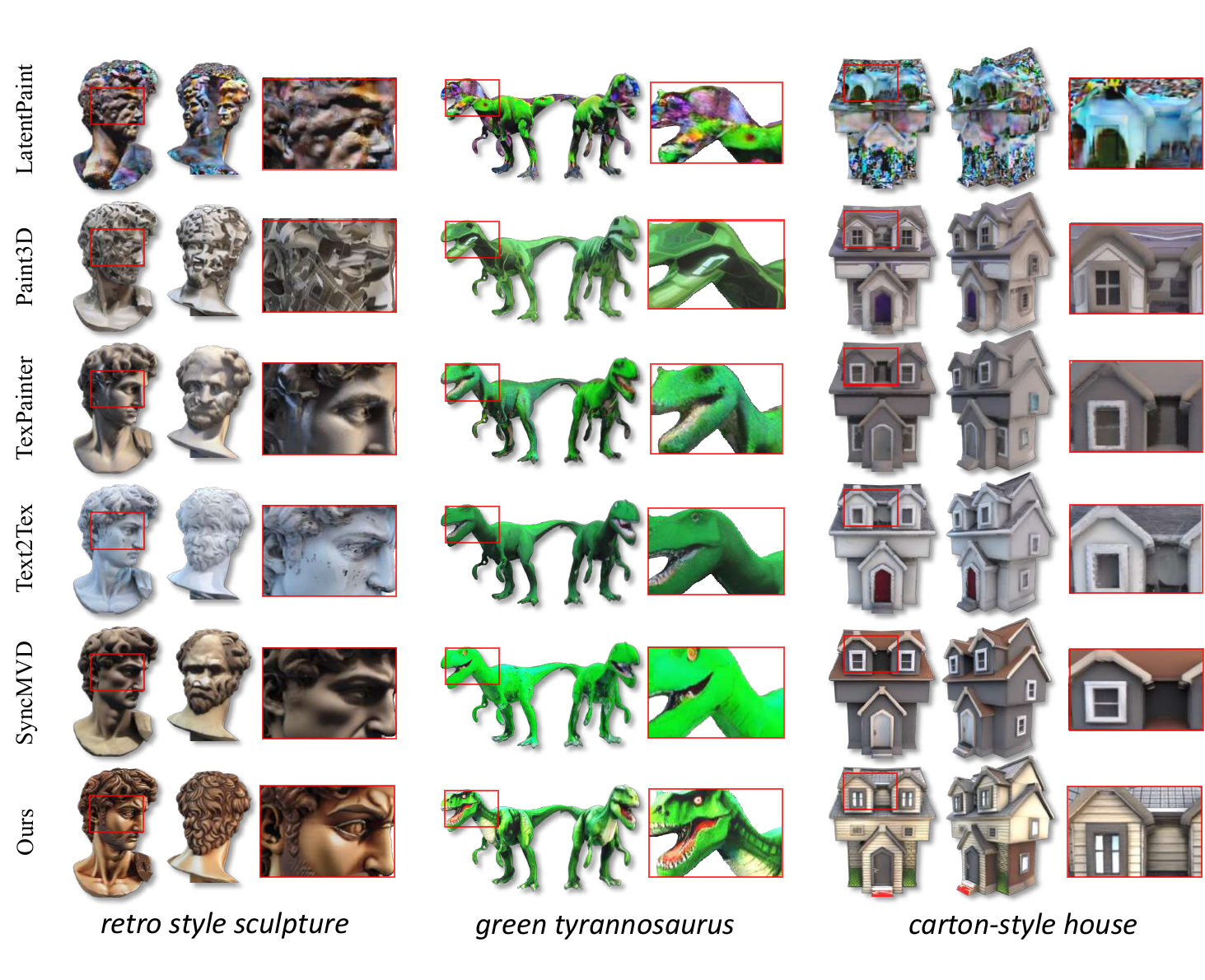} 
\caption{Qualitative comparisons on texture generation conditioned on text prompt. 
% We compare against Latent-Paint~\cite{luo2023latent}, Paint3D~\cite{zeng2024paint3d}, TexPainter~\cite{zhang2024texpainter}, Text2Tex~\cite{chen2023text2tex} and SyncMVD~\cite{liu2023text}. 
While previous methods result in an over-smoothed appearance and global inconsistencies such as the Janus problem, our method generates globally consistent and higher-quality textures rich in detail (refer to supplements for more results).}
\label{fig: t2t_comparison}
\end{figure*}

\section{Experiments}
% In this section, we demonstrate the effectiveness of our proposed methods quantitatively and qualitatively. We first provide the implementation details, dataset, evaluation metrics, and baselines in Sec.~\ref{sec: setup}. Then provide quantitative and qualitative analysis in Sec.~\ref{sec: analysis_quan} and Sec.~\ref{sec: analysis_qual}. We also provide the ablation studies to single out the improvement of each proposed module in Sec.~\ref{sec: ablation}. 

\subsection{Setup}
\label{sec: setup}
% We introduce the implementation details, dataset, evaluation metrics, and baselines in this section.

\noindent\textbf{Implementation Details.} Our experiments are conducted on an NVIDIA A100 GPU. 
% We utilize the official Stable Diffusion XL model with the Controlnet(depth-sdxl-1.0) and IP-Adaptor finetuned on SD XL. 
For denoising epoch, we use DDIM~\cite{song2020denoising} as the sampler. We set the number of iterations to 30 steps, the CFG scale (classifier-free guidance scale) for Direction-Aware Adaptation to 12, and the scale of Visual Guidance Enhancement to 0.6. Texture warping for latent views is used in the first 24 steps. 
We sample 8 views for a mesh, and the elevations and azimuths are ($-180^{\circ}$, $15^{\circ}$), ($-120^{\circ}$, $-15^{\circ}$), ($-60^{\circ}$, $15^{\circ}$), ($0^{\circ}$, $-15^{\circ}$), ($60^{\circ}$, $15^{\circ}$), ($120^{\circ}$, $-15^{\circ}$), ($-180^{\circ}$, $-45^{\circ}$), ($0^{\circ}$, $45^{\circ}$).
We implement the rendering function by Pytorch3D~\cite{ravi2020pytorch3d, paszke2017automatic}.
% for depth maps, latent feature maps, and RGB maps. 

\noindent\textbf{Dataset. } We collect 60 meshes with corresponding text prompts and 60 meshes with corresponding image prompts to evaluate the text-to-texture and image-to-texture generation ability, respectively. These meshed are randomly sampled from Objaverse~\cite{objaverse}, Objaverse-XL~\cite{objaverseXL} and ShapeNet~\cite{chang2015shapenet}.

\noindent\textbf{Evaluation metrics.} Fréchet inception distance (FID) and Kernel Inception Distance (KID) measure the feature dissimilarity between two image collections, with feature extraction performed using the Inception V3~\cite{szegedy2016rethinking}, as metrics for the realism and diversity. Moreover, we also utilize the CLIP Score metric~\cite{taited2023CLIPScore} to assess the alignment with original inputs.

\noindent\textbf{Baselines. } For text-to-texture generation, we compare \name with inpainting-based methods \textbf{Text2Tex}~\cite{chen2023text2tex}, \textbf{Paint3D}~\cite{zeng2024paint3d}, optimization-based methods \textbf{Latent-Paint}~\cite{metzer2023latent}, synchronization-based methods \textbf{TexPainter}~\cite{zhang2024texpainter}, \textbf{SyncMVD}~\cite{liu2023text}. For image-to-texture generation, we compare our~\name with inpainting-based methods \textbf{TEXTure}~\cite{richardson2023texture}, \textbf{Paint3D}~\cite{zeng2024paint3d}, optimization-based methods \textbf{PGC-3D}~\cite{pan2024enhancing}. For a fair comparison, we re-run their official codes using the same meshes and prompts.

\begin{figure*}[tbp]
\centering
\includegraphics[width=0.9\textwidth, trim=0 0 10 10, clip]{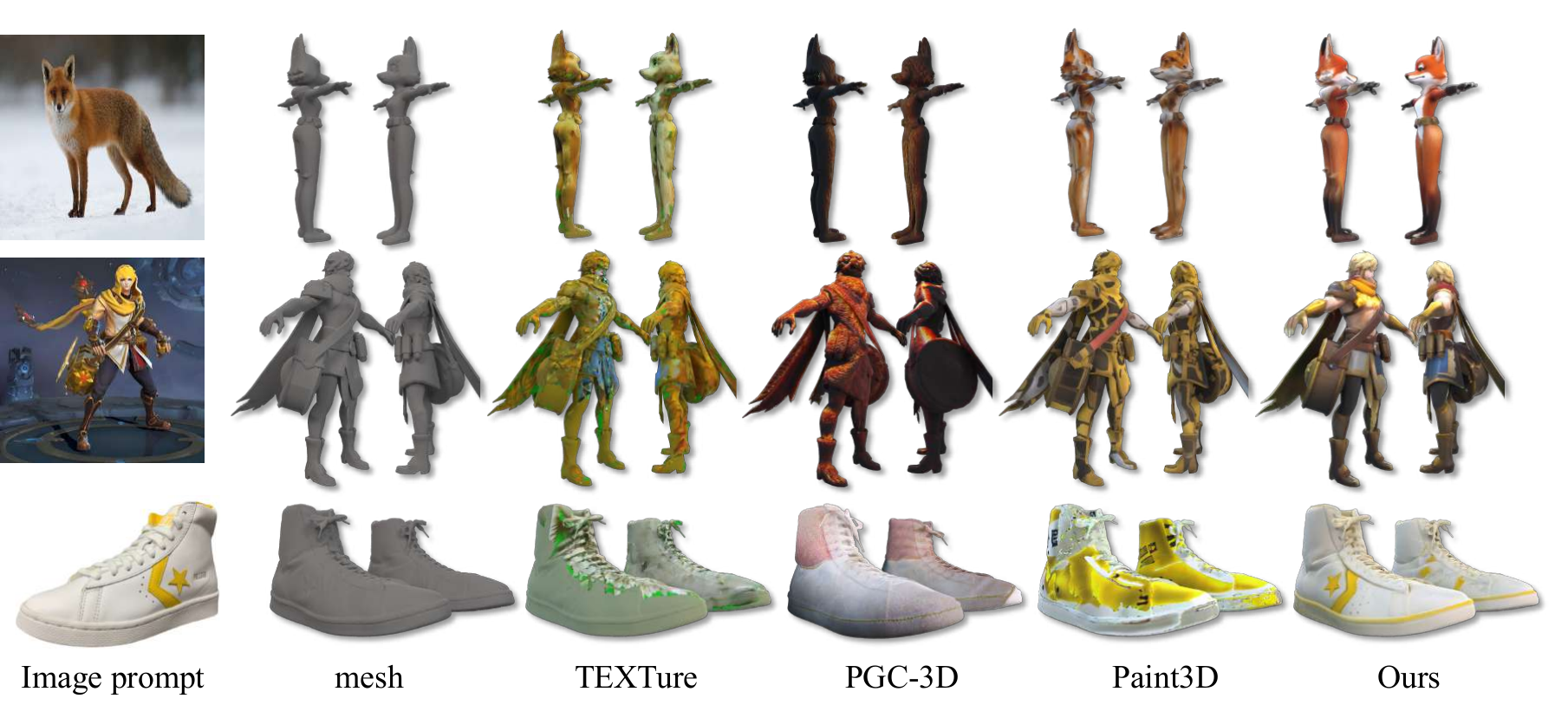} 
\caption{Qualitative comparisons on texture generation conditioned on image prompt. Compared with baselines, \name{} achieves better alignment with input images on target meshes. Our results are also superior in texture integrity and quality.
% We compare against TEXTure~\cite{yu2023texture}, PGC-3D~\cite{pan2024enhancing} and Paint3D~\cite{zeng2024paint3d}, and achieve better alignment with input images on target meshes. Our results are also superior in terms of texture integrity and quality. 
}
\label{fig: i2t_comparison}
\end{figure*}

\subsection{Quantitative Analysis}
\label{sec: analysis_quan}
\noindent\textbf{FID \& KID. } 
% Fréchet inception distance (FID) and Kernel Inception Distance (KID) measure the feature dissimilarity between two image collections, with feature extraction performed using the Inception V3~\cite{szegedy2016rethinking}, as metrics for the realism and diversity.
Following GenesisTex~\cite{gao2024genesistex}, we use samples from pre-trained diffusion models as ground truth labels. We render depth maps from 16 cameras around the mesh as ControlNet inputs and we maintain the same prompts for inference. We modify the ground truth images' background to white to ensure the focus remains on the texture. We render each mesh from the same views, assessing quality via FID and KID. Tab.~\ref{tab: t2t_comparison} shows our method outperforms baselines, indicating superior synthesis quality and closer alignment to ground truth, attributed to the explicit information provided by visual guidance.

\noindent\textbf{ClipScore. } 
% We utilize the CLIP Score metric~\cite{taited2023CLIPScore} to assess the alignment of generated results with original inputs. 
For text-to-texture tasks, the CLIP score is derived by comparing rendered views with text prompts, while for image-to-texture tasks, it is computed by comparing rendered views with image prompts. Semantic consistency deviations may occur in text-to-texture tasks due to the text-to-image module, resulting in a slightly lower CLIP Score than TexPainter, as shown in Tab.~\ref{tab: t2t_comparison}. However, our method enhances texture quality and realism. For image-to-texture tasks, Paint3D improves alignment by injecting image features into UV map refinement, but struggles with complex UV maps due to semantic differences. Conversely, \name{} achieves the highest CLIP Score by employing visual guidance on multi-view inference, ensuring semantic consistency with input images.

\noindent\textbf{Speed. } Compared with optimization-based and inpainting-based methods requiring sequential sampling, our approach simultaneously generates multiple views once. Rapid texture warping further accelerates generation, in contrast to TexPainter using 40-minute due to enforced differentiable rendering in each denoising step.

\begin{table}[t!]
\centering
\tabcolsep=0.12cm
\subtable[Quantitative comparisons conditioned on text prompt.]{
\begin{tabular}{c c c c c}

\toprule
    & FID $\downarrow$ & KID $\downarrow$  & CLIP Score $\uparrow$ & Time $\downarrow$ \\
\midrule
Latent-Paint   & 137.56       & 44.09  & 31.32 & 8 min\\
Paint3D        & 99.05       & 13.28   & 33.44 & 3.1 min\\
TexPainter     & 96.97       & 13.51   & \textbf{34.48} & 40 min\\
Text2Tex       & 	90.98      & 10.62 & 34.33 & 6 min\\
SyncMVD        &   87.65      & 9.53   &  34.30 & \textbf{1.8 min}\\
Ours           & \textbf{81.72}      & \textbf{8.78} & 34.31 & 2.2 min\\
\bottomrule
\end{tabular}
\label{firsttable}
}
 
\qquad
 
\subtable[Quantitative comparisons conditioned on image prompt.]{        
\begin{tabular}{c c c c c}
\toprule
    & FID $\downarrow$ & KID $\downarrow$  & CLIP Score $\uparrow$ & Time $\downarrow$ \\
\midrule
TEXTure         & 141.26       & 51.11 & 82.92 & 32 min\\
PGC-3D          & 133.76	   & 32.37 & 79.05 & 19 min\\
Paint3D         & 103.44       & 18.71 &  87.43 & 3.2 min\\
Ours            & \textbf{82.52}    & \textbf{11.97} & \textbf{89.99} &  \textbf{2.4 min}\\
\bottomrule
\end{tabular}
\label{secondtable}
}
\caption{Quantitative comparisons with baselines. 
$\downarrow$ and $\uparrow$ are used to indicate the performance in relation to the score. 
$\downarrow$ indicates better performance with a lower score, while $\uparrow$ indicates better performance with a higher score.
}
\label{tab: t2t_comparison}
\end{table}

\subsection{Qualitative Analysis}
\label{sec: analysis_qual}
As presented in Fig.~\ref{fig: t2t_comparison} and Fig.~\ref{fig: i2t_comparison},
we qualitatively evaluate the quality of the generated texture conditioned on text and image, respectively. The Latent-Paint method produces textures that are broken by noise and impurities, as a consequence of the inherent limitation of SDS. Both Paint3D and Text2Tex exhibit noticeable blurriness at texture seams and suffer from the multi-face issue, thereby failing to maintain multi-view consistency. TexPainter and SyncMVD struggle to preserve clarity on texture, ending up with an over-smoothed and monotonous appearance. Through Visual Guidance Enhancement, \name{} overcomes the over-smooth problem and preserves more high-frequency details during generation. Direction-Aware adaptation also strengthens geometric alignment on side or back views, alleviating the Janus problem.

For image-to-texture tasks shown in Fig.~\ref{fig: i2t_comparison}, both TEXTure and PGC-3D struggle to retain the semantic information from image prompts, resulting in low-quality, noisy textures. Paint3D, while partially preserving semantic information, generates a significant number of texture fragments. In contrast, we implement a decoupled cross-attention strategy from visual features and direction features, achieving semantic alignment with the style of image prompts on target meshes and exhibiting a vibrant appearance.

\subsection{Ablation Studies}
\label{sec: ablation}
\begin{figure}
    \centering
    \includegraphics[width=1\linewidth, trim=0 0 0 30, clip ]{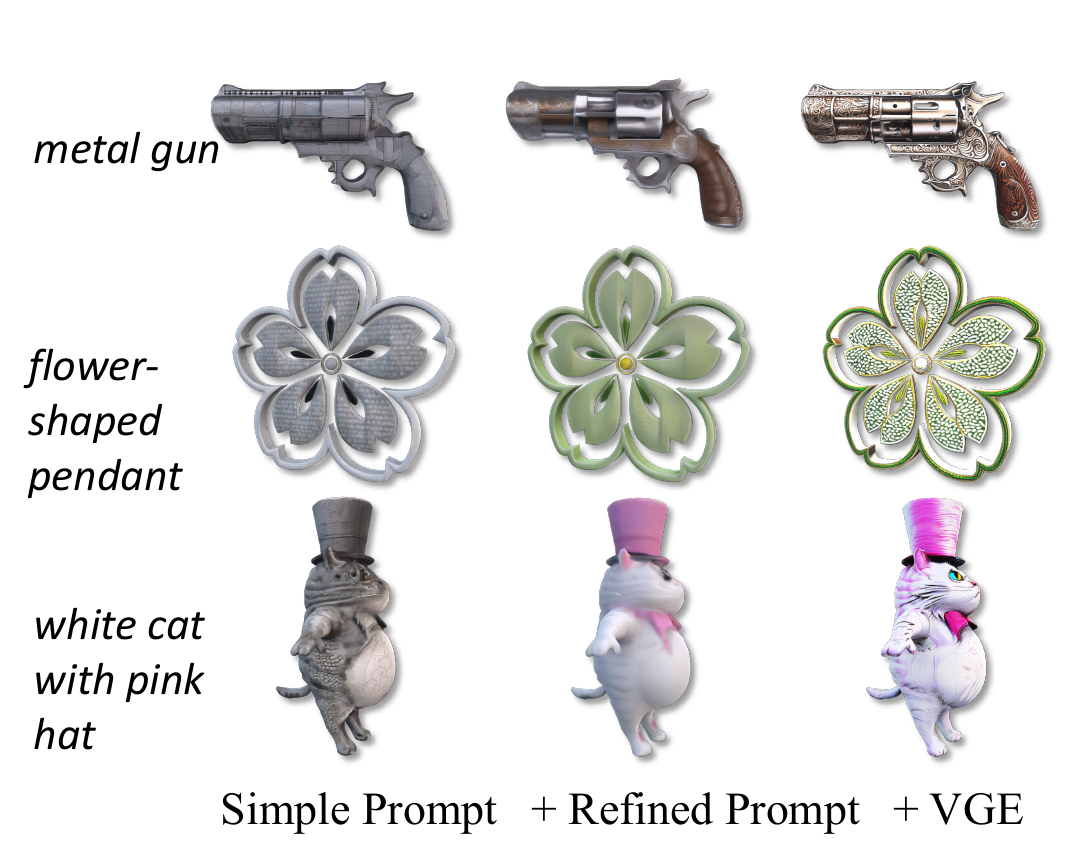}
    \caption{Ablation results on Visual Guidance Enhancement. 
    % Compared with textual guidance, Visual Guidance Enhancement improves texture richness greatly. 
    }
    \label{fig: abla_enhance}
\end{figure}

\begin{figure}
    \centering
    \includegraphics[width=0.95\linewidth, trim=10 0 0 10, clip ]{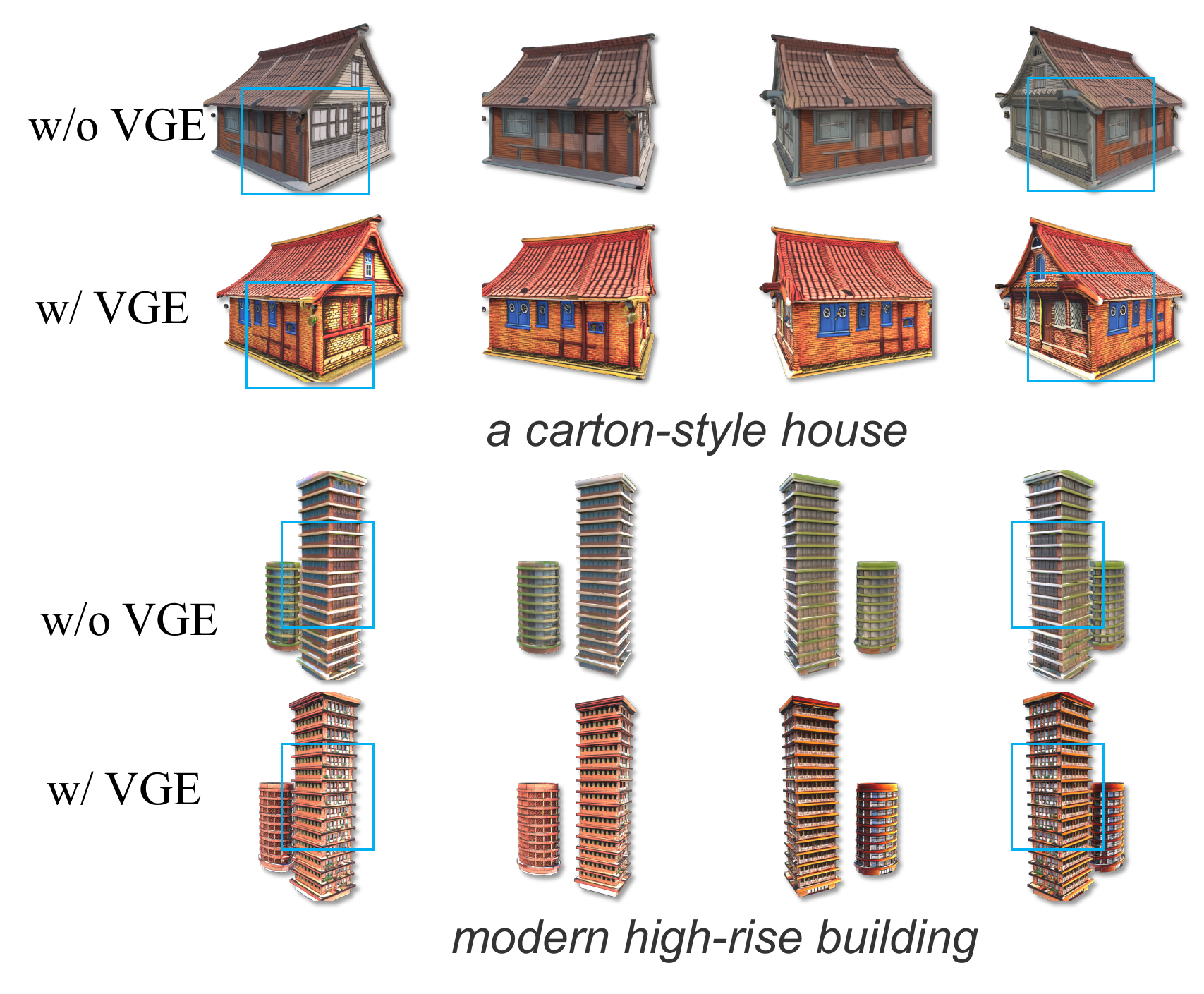}
    \caption{Ablation results on Visual Guidance Enhancement. 
    % Visual Guidance Enhancement maintains consistent color and improves global style consistency on each side. 
    }
    \label{fig: abla_style}
\end{figure}

\begin{figure}
    \centering
    \includegraphics[width=0.8\linewidth, trim=10 0 0 0, clip ]{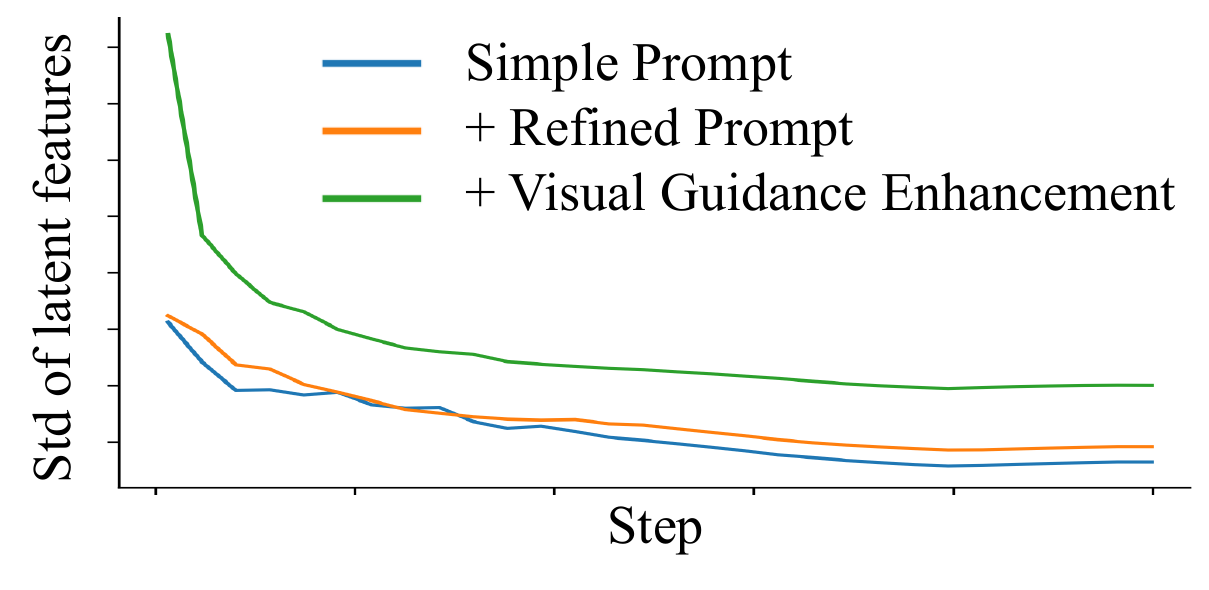}
    \caption{Ablation results of standard deviation (std) on Visual Guidance Enhancement. 
    % Using refined long prompts can slightly improve std, while with Visual Guidance Enhancement, we achieve the highest std.
    }
    \label{tab: abla_enhance_std}
\end{figure}

\noindent\textbf{Effectiveness of Visual Guidance Enhancement. }
To examine the influence of various prompts on texture quality, we compare three types: \textbf{simple prompts} comprising no more than five words, \textbf{refined prompts} expanded by a Large Language Model (LLM) Llama-3~\cite{llama3modelcard}, and \textbf{Visual Guidance Enhancement} using our method. As depicted in Fig.~\ref{fig: abla_enhance}, textures derived from simple prompts can appear blurry or desaturated. While refined prompts marginally slightly address this issue, their results still lack vibrant details. However, with Visual Guidance Enhancement, we can observe a significant improvement in detail richness. 
Similar to VCD-Texture~\cite{liu2024vcd}, we visualize the standard deviation (std) curve of the latent features from foreground parts during the denoising phase, in Table~\ref{tab: abla_enhance_std}. Here, Visual Guidance Enhancement achieves the highest feature variance, indicating high-frequency details. Apart from the rich details brought by VGA, we also achieve better \textbf{style consistency} in Fig.~\ref{fig: abla_style}. For example, for prompt \textit{'a carton-style house'}, the original results can show grey on one side and orange on the other side, while with the VGA module, these two sides show consistent color.

\begin{figure}
    \centering
    \includegraphics[width=\linewidth, trim=0 70 80 50, clip ]{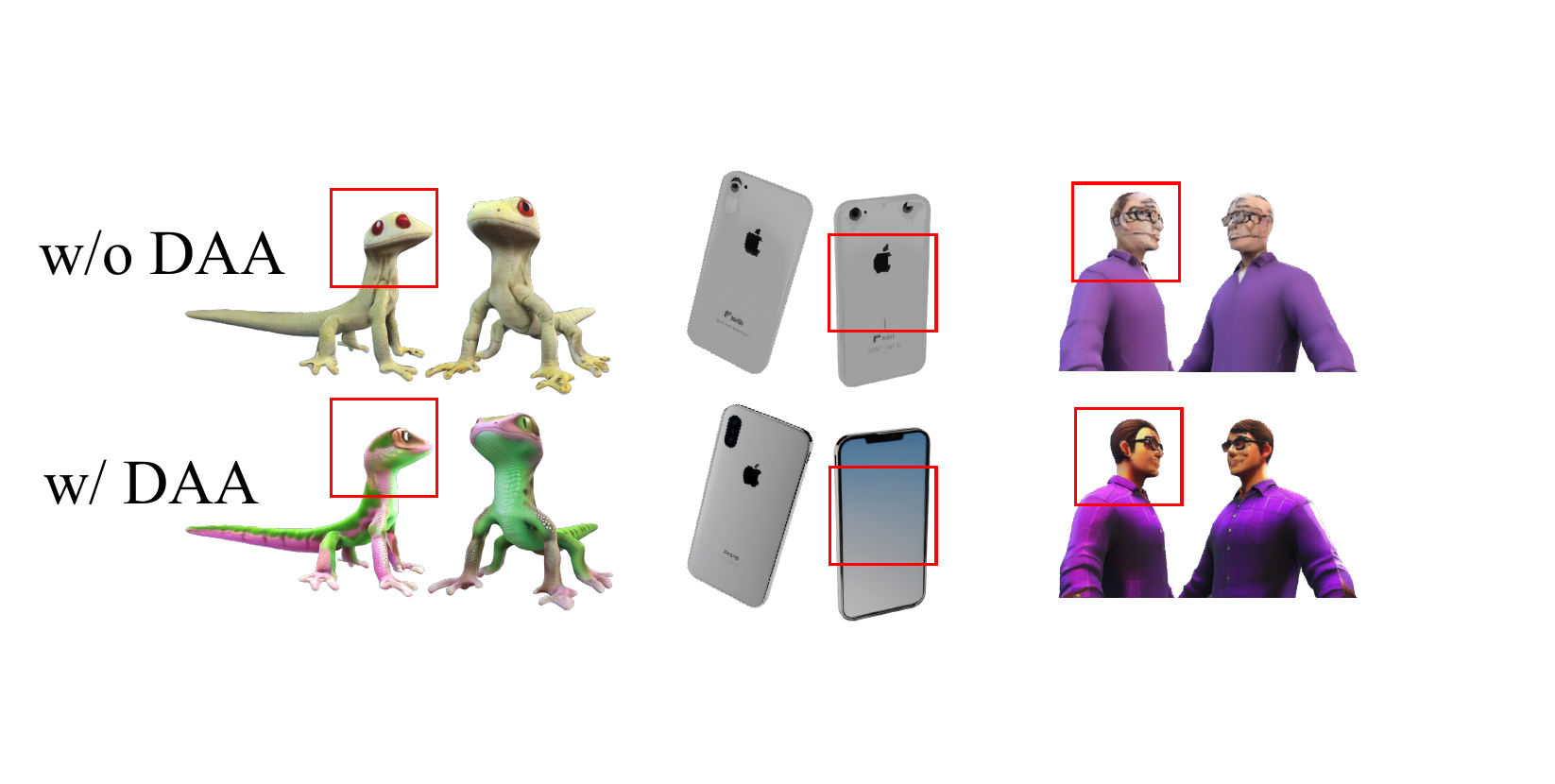}
    \caption{Ablation results on Direction-Aware Adaptation. We show results from \textbf{side} and \textbf{front} views.}
    \label{fig: vaa}
\end{figure}

\noindent\textbf{Effectiveness of Direction-Aware Adaptation. } We visualize the effects of Direction-Aware Adaptation (DAA) on alleviating multi-face problem in Fig.~\ref{fig: vaa}. With reinforcement of specific direction on target meshes, the contents of each view are correct(e.g., a lizard shows one eye from the side view, the front of the iPhone is a screen, and a human only has one face). We also experiment on 200 human cases, 200 animal cases, and 200 object cases. Tab.~\ref{tab: vaa} shows a lower multi-face percentage.

\section{Conclusion}
In this paper, we introduce FlexiTex, a novel framework designed for high-fidelity texture generation on 3D objects, accommodating both text and image prompts. We also mitigate multi-face issues across various types of objects. Experiments demonstrate that \name{} is superior to baseline methods in both quantitative and qualitative measures.
However, our method does have certain limitations. Specifically, our generated results have not yet decoupled lighting information, leading to potential artifacts in highlights or shadows. Additionally, in areas where multiple views are inconsistent or unobserved, minor dirty spots may appear. Future research could explore material generation for re-lighting tasks and improved texture warping strategies, such as maintaining a color field for smooth transitions in conflict areas.

\begin{table}
\centering
\begin{tabular}{c c c c}
\toprule
     & Animal $\downarrow$ & Object $\downarrow$& Human $\downarrow$\\
\midrule
w/o DAA      & 48.00 \% & 8.00 \% & 24.00 \%\\
w/ DAA       & \textbf{21.50 \%} & \textbf{3.00 \%} & \textbf{16.50 \%}\\
\bottomrule
\end{tabular}
\caption{Ablation results on Direction-Aware Adaptation (DAA). We calculate multi-face percentages on 200 human cases, 200 animal cases, and 200 object cases.}
\label{tab: vaa}
\end{table}

\section{Acknowledgments}
This work is support in part by the National Natural Science Foundation of China (No. 62072330), and in part by Tianjin Science and Technology Plan Project (No. 23ZYCGCG00710).

\bibliography{aaai25}

\end{document}